\documentclass[11pt,letterpaper]{article}
\usepackage{authblk}
\usepackage{emnlp2016}
\usepackage{times}
\usepackage{latexsym}
\usepackage{tikz-qtree}
\usepackage{amssymb}
\usepackage{amsmath}
\usepackage{multirow}
\usepackage[utf8]{inputenc}
\usepackage[T1]{fontenc}
\usepackage{graphicx}

\makeatletter
\newcommand{\@BIBLABEL}{\@emptybiblabel}
\newcommand{\@emptybiblabel}[1]{}
\makeatother
\usepackage[hidelinks]{hyperref}

\emnlpfinalcopy

\def\emnlppaperid{499}

\title{A Hierarchical Model of Reviews for Aspect-based Sentiment Analysis}

\author[1,2]{Sebastian Ruder}
\author[2]{Parsa Ghaffari}
\author[1]{John G. Breslin}
\affil[1]{Insight Centre for Data Analytics}
\affil[ ]{National University of Ireland, Galway}
\affil[ ]{\tt \{sebastian.ruder,john.breslin\}@insight-centre.org}
\affil[2]{Aylien Ltd.}
\affil[ ]{Dublin, Ireland}
\affil[ ]{\tt \{sebastian,parsa\}@aylien.com}

\date{}

\begin{document}

\maketitle
\begin{abstract}
Opinion mining from customer reviews has become pervasive in recent years. Sentences in reviews, however, are usually classified independently, even though they form part of a review's argumentative structure. Intuitively, sentences in a review build and elaborate upon each other; knowledge of the review structure and sentential context should thus inform the classification of each sentence. We demonstrate this hypothesis for the task of aspect-based sentiment analysis by modeling the interdependencies of sentences in a review with a hierarchical bidirectional LSTM. We show that the hierarchical model outperforms two non-hierarchical baselines, obtains results competitive with the state-of-the-art, and outperforms the state-of-the-art on five multilingual, multi-domain datasets without any hand-engineered features or external resources.
\end{abstract}

\section{Introduction}

Sentiment analysis \cite{Pang2008} is used to gauge public opinion towards products, to analyze customer satisfaction, and to detect trends. With the proliferation of customer reviews, more fine-grained aspect-based sentiment analysis (ABSA) has gained in popularity, as it allows aspects of a product or service to be examined in more detail.

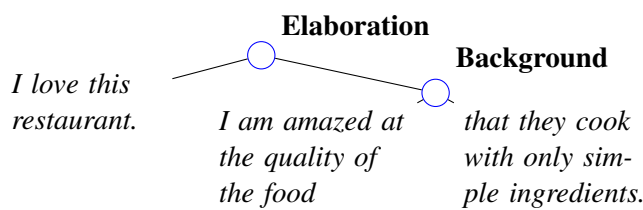
\begin{figure}
\centering
\begin{tikzpicture}[
   level distance=0.5cm,sibling distance=.5cm, 
   edge from parent path={(\tikzparentnode) -- (\tikzchildnode)}]
\Tree [. \node[blue, style={draw,circle}, label=above right:\textbf{Elaboration}](parentnode){}; 
    [.\node [text width=2cm] {\textit{I love this restaurant.}};]  
    [.\node [blue, style={draw,circle}, label=above right:\textbf{Background}] {};
       [.\node [text width=2.5cm] {\textit{I am amazed at the quality of the food}};]
       [.\node [text width=2.5cm] {\textit{that they cook with only simple ingredients.}};]
    ] ]
\end{tikzpicture}
\caption{RST structure of an example review.}
\label{fig:rst_structure}
\end{figure}

Reviews -- just with any coherent text -- have an underlying structure. A visualization of the discourse structure according to Rhetorical Structure Theory (RST) \cite{Mann1988} for the example review in Figure \ref{fig:rst_structure} reveals that sentences and clauses are connected via different rhetorical relations, such as \textit{Elaboration} and \textit{Background}.

Intuitively, knowledge about the relations and the sentiment of surrounding sentences should inform the sentiment of the current sentence. If a reviewer of a restaurant has shown a positive sentiment towards the quality of the food, it is likely that his opinion will not change drastically over the course of the review. Additionally, overwhelmingly positive or negative sentences in the review help to disambiguate sentences whose sentiment is equivocal.

Neural network-based architectures that have recently become popular for sentiment analysis and ABSA, such as convolutional neural networks \cite{Severyn2015a}, LSTMs \cite{Vo2015}, and recursive neural networks \cite{Nguyen2015a}, however, are only able to consider intra-sentence relations such as \textit{Background} in Figure \ref{fig:rst_structure} and fail to capture inter-sentence relations, e.g. \textit{Elaboration} that rely on discourse structure and provide valuable clues for sentiment prediction.

We introduce a hierarchical bidirectional long short-term memory (H-LSTM) that is able to leverage both intra- and inter-sentence relations. The sole dependence on sentences and their structure within a review renders our model fully language-independent. We show that the hierarchical model outperforms strong sentence-level baselines for aspect-based sentiment analysis, while achieving results competitive with the state-of-the-art and outperforming it on several datasets without relying on any hand-engineered features or sentiment lexica.

\section{Related Work}

\textbf{Aspect-based sentiment analysis.} Past approaches use classifiers with expensive hand-crafted features based on n-grams, parts-of-speech, negation words, and sentiment lexica \cite{Pontiki2014a,Pontiki2015}. The model by Zhang and Lan \shortcite{Zhang2015f} is the only approach we are aware of that considers more than one sentence. However, it is less expressive than ours, as it only extracts features from the preceding and subsequent sentence without any notion of structure. Neural network-based approaches include an LSTM that determines sentiment towards a target word based on its position \cite{Tang2015} as well as a recursive neural network that requires parse trees \cite{Nguyen2015a}. In contrast, our model requires no feature engineering, no positional information, and no parser outputs, which are often unavailable for low-resource languages. We are also the first -- to our knowledge -- to frame sentiment analysis as a sequence tagging task.

\textbf{Hierarchical models.} Hierarchical models have been used predominantly for representation learning and generation of paragraphs and documents: Li et al. \shortcite{Li2015d} use a hierarchical LSTM-based autoencoder to reconstruct reviews and paragraphs of Wikipedia articles. Serban et al. \shortcite{Serban2016} use a hierarchical recurrent encoder-decoder with latent variables for dialogue generation. Denil et al. \shortcite{Denil2014} use a hierarchical ConvNet to extract salient sentences from reviews, while Kotzias et al. \shortcite{Kotzias2015} use the same architecture to learn sentence-level labels from review-level labels using a novel cost function. The model of Lee and Dernoncourt \shortcite{Lee2016} is perhaps the most similar to ours. While they also use a sentence-level LSTM, their class-level feed-forward neural network is only able to consider a limited number of preceding texts, while our review-level bidirectional LSTM is (theoretically) able to consider an unlimited number of preceding \textit{and} successive sentences.

\section{Model}

\begin{figure*}
	\centering
  	\includegraphics[width=0.75\linewidth]{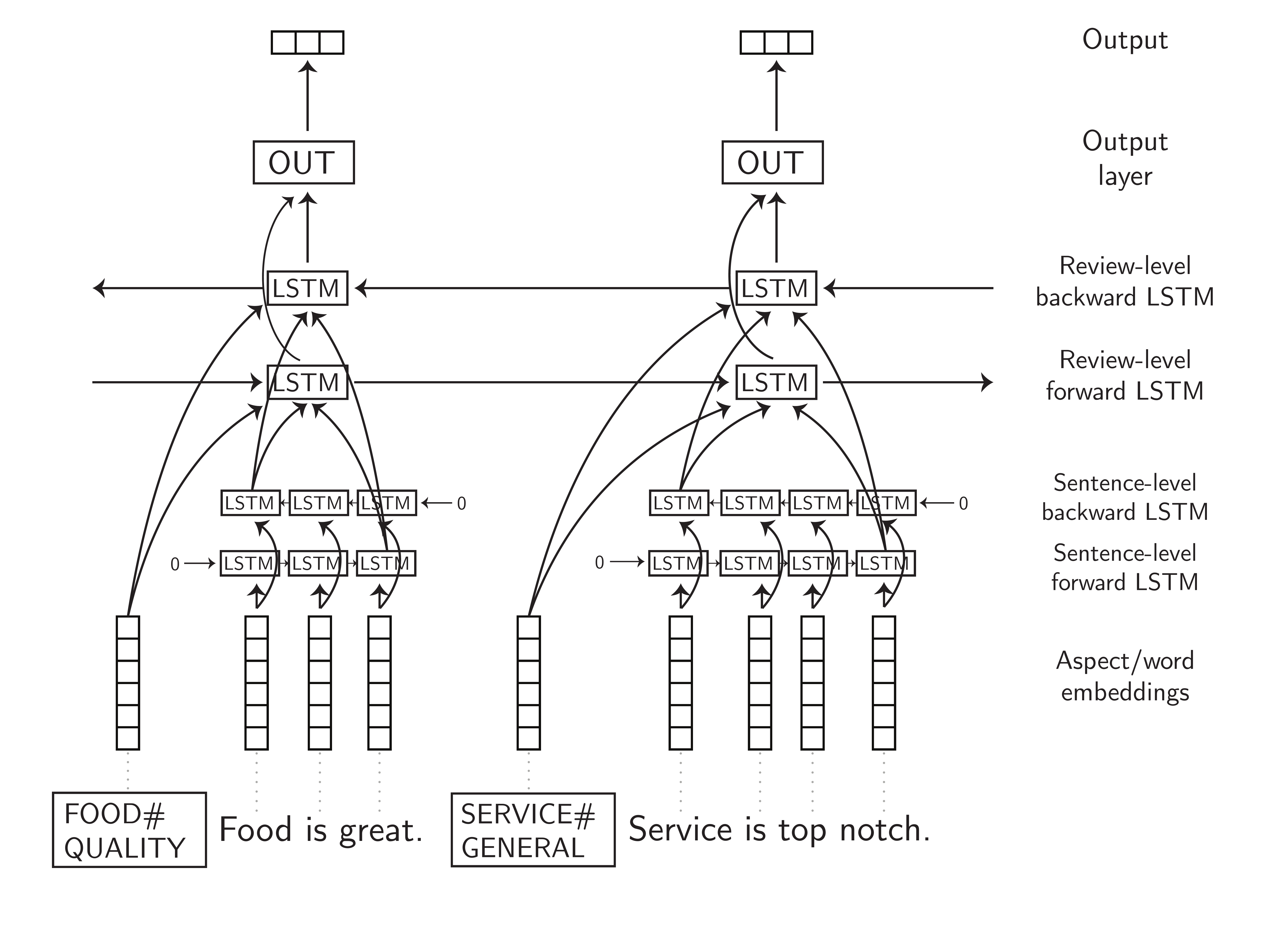}
  	\caption{The hierarchical bidirectional LSTM (H-LSTM) for aspect-based sentiment analysis. Word embeddings are fed into a sentence-level bidirectional LSTM. Final states of forward and backward LSTM are concatenated together with the aspect embedding and fed into a bidirectional review-level LSTM. At every time step, the output of the forward and backward LSTM is concatenated and fed into a final layer, which outputs a probability distribution over sentiments.}
  	\label{fig:hierarchical_lstm}
\end{figure*}

In the following, we will introduce the different components of our hierarchical bidirectional LSTM architecture displayed in Figure \ref{fig:hierarchical_lstm}.

\subsection{Sentence and Aspect Representation}

Each review consists of sentences, which are padded to length $l$ by inserting padding tokens. Each review in turn is padded to length $h$ by inserting sentences containing only padding tokens. We represent each sentence as a concatentation of its word embeddings $x_{1:l}$ where $x_t \in \mathbb{R}^k$ is the $k$-dimensional vector of the $t$-th word in the sentence.

Every sentence is associated with an aspect. Aspects consist of an entity and an attribute, e.g. \texttt{FOOD\#QUALITY}. Similarly to the entity representation of Socher et al. \shortcite{Socher2013}, we represent every aspect $a$ as the average of its entity and attribute embeddings $\frac{1}{2} (x_e + x_a) $ where $x_e, x_a \in \mathbb{R}^m$ are the $m$-dimensional entity and attribute embeddings respectively\footnote{Averaging embeddings produced slightly better results than using a separate embedding for every aspect.}.

\subsection{LSTM}

We use a Long Short-Term Memory (LSTM) \cite{Hochreiter1997}, which adds input, output, and forget gates to a recurrent cell, which allow it to model long-range dependencies that are essential for capturing sentiment.

For the $t$-th word in a sentence, the LSTM takes as input the word embedding $x_t$, the previous output $h_{t-1}$ and cell state $c_{t-1}$ and computes the next output $h_t$ and cell state $c_t$. 
% equations are numbered without *
%\begin{align*}
%i_t & = \sigma(W_i x_t + U_i h_{t-1} + b_i) \\
%f_t & = \sigma(W_f x_t + U_f h_{t-1} + b_f) \\
%c_t & = f_t \odot c_{t-1} + i_t \odot \text{tanh}(W_c x_t + U_c h_{t-1} + b_c) \\
%o_t & = \sigma(W_o x_t + U_o h_{t-1} + b_o) \\
%h_t & = o_t \odot \text{tanh}(c_t)
%\end{align*}
%
%where $i_t$, $f_t$, and $o_t$ are the input, forget, and output gates respectively, $W_j \in \mathbb{R}^{k \times m}$, $U_j \in \mathbb{R}^{k \times k}$ are weight matrices and $b_j \in \mathbb{R}^k$ are bias vectors for $j \in {i, f, c, o}$, $\text{tanh}$ and $\sigma$ denote the element-wise application of the hyperbolic tangent and sigmoid function respectively and $\odot$ designates the element-wise multiplication of two vectors.
Both $h$ and $c$ are initialized with zeros.

\subsection{Bidirectional LSTM}

Both on the review and on the sentence level, sentiment is dependent not only on preceding but also successive words and sentences. A Bidirectional LSTM (Bi-LSTM) \cite{Graves2013a} allows us to look ahead by employing a forward LSTM, which processes the sequence in chronological order, and a backward LSTM, which processes the sequence in reverse order. The output $h_t$ at a given time step is then the concatenation of the corresponding states of the forward and backward LSTM.

\subsection{Hierarchical Bidirectional LSTM}

Stacking a Bi-LSTM on the review level on top of sentence-level Bi-LSTMs yields the hierarchical bidirectional LSTM (H-LSTM) in Figure \ref{fig:hierarchical_lstm}.

The sentence-level forward and backward LSTMs receive the sentence starting with the first and last word embedding $x_{1}$ and $x_l$ respectively. The final output $h_l$ of both LSTMs is then concatenated with the aspect vector $a$\footnote{We experimented with other interactions, e.g. rescaling the word embeddings by their aspect similarity, an attention-like mechanism, as well as summing and multiplication, but found that simple concatenation produced the best results.} and fed as input into the review-level forward and backward LSTMs. The outputs of both LSTMs are concatenated and fed into a final softmax layer, which outputs a probability distribution over sentiments\footnote{The sentiment classes are \textit{positive}, \textit{negative}, and \textit{neutral}.} for each sentence.

\section{Experiments}

\subsection{Datasets}

For our experiments, we consider datasets in five domains (restaurants, hotels, laptops, phones, cameras) and eight languages (English, Spanish, French, Russian, Dutch, Turkish, Arabic, Chinese) from the recent SemEval-2016 Aspect-based Sentiment Analysis task \cite{SemEval2016:task5}, using the provided train/test splits. In total, there are 11 domain-language datasets containing 300-400 reviews with 1250-6000 sentences\footnote{Exact dataset statistics can be seen in \cite{SemEval2016:task5}.}. Each sentence is annotated with none, one, or multiple domain-specific aspects and a sentiment value for each aspect.

\begin{table*}[]
\centering
\begin{tabular}{l l c c c c  c c c}
\textbf{Language} & \textbf{Domain} & \texttt{Best} & \texttt{XRCE} & \texttt{IIT-TUDA} & \texttt{CNN} & \texttt{LSTM} & \texttt{H-LSTM} & \texttt{HP-LSTM} \\\hline
English & Restaurants &  \textbf{88.1} & \textbf{88.1} & 86.7 & 82.1 & 81.4 & 83.0 & 85.3 \\
Spanish & Restaurants & \textbf{83.6} & - & \textbf{83.6} & 79.6 & 75.7 & 79.5 & 81.8 \\
French & Restaurants & \textbf{78.8} & \textbf{78.8} & 72.2 & 73.2 & 69.8 & 73.6 & 75.4 \\ 
Russian & Restaurants & 77.9 & - &73.6 & 75.1 & 73.9 & \textbf{78.1} & 77.4 \\
Dutch & Restaurants & 77.8 & - & 77.0 & 75.0 & 73.6 & 82.2 & \textbf{84.8} \\
Turkish & Restaurants & \textbf{84.3} & - & \textbf{84.3} & 74.2 & 73.6 & 76.7 & 79.2\\ 
Arabic & Hotels & 82.7 & - & 81.7 & 82.7 & 80.5 & 82.8 & \textbf{82.9} \\
English & Laptops & \textbf{82.8} & - & \textbf{82.8} & 78.4 & 76.0 & 77.4 & 80.1 \\
Dutch & Phones & 83.3 & - & 82.6 & 83.3 & 81.8 & 81.3 & \textbf{83.6}\\
Chinese & Cameras & \textbf{80.5} & - & - & 78.2 & 77.6 & 78.6 & 78.8 \\
Chinese & Phones & 73.3 & - & - & 72.4 & 70.3 & \textbf{74.1} & 73.3
\end{tabular}
\caption{Results of our system with randomly initialized word embeddings (\texttt{H-LSTM)} and with pre-trained embeddings (\texttt{HP-LSTM}) for ABSA for each language and domain in comparison to the best system for each pair (\texttt{Best}), the best two single systems (\texttt{XRCE}, \texttt{IIT-TUDA}), a sentence-level CNN (\texttt{CNN}), and our sentence-level LSTM (\texttt{LSTM}).}
\label{tab:results}
\end{table*}

\subsection{Training Details}

Our LSTMs have one layer and an output size of 200 dimensions. We use 300-dimensional word embeddings. We use pre-trained GloVe \cite{Pennington2014} embeddings for English, while we train embeddings on \texttt{frWaC}\footnote{\url{http://wacky.sslmit.unibo.it/doku.php?id=corpora}} for French and on the Leipzig Corpora Collection\footnote{\url{http://corpora2.informatik.uni-leipzig.de/download.html}} for all other languages.\footnote{Using 64-dimensional Polyglot embeddings \cite{Al-Rfou2013} yielded generally worse performance.} Entity and attribute embeddings of aspects have 15 dimensions and are initialized randomly. We use dropout of 0.5 after the embedding layer and after LSTM cells, a gradient clipping norm of 5, and no $l_2$ regularization.

We unroll the aspects of every sentence in the review, e.g. a sentence with two aspects occurs twice in succession, once with each aspect. We remove sentences with no aspect\footnote{Labeling them with a \texttt{NONE} aspect and predicting \textit{neutral} slightly decreased performance.} and ignore predictions for all sentences that have been added as padding to a review so as not to force our model to learn meaningless predictions, as is commonly done in sequence-to-sequence learning \cite{Sutskever2014}. We segment Chinese data before tokenization.

We train our model to minimize the cross-entropy loss, using stochastic gradient descent, the Adam update rule \cite{Kingma2015}, mini-batches of size 10, and early stopping with a patience of 10.

\subsection{Comparison models}

We compare our model using random (\texttt{H-LSTM}) and pre-trained word embeddings (\texttt{HP-LSTM})  against the best model of the SemEval-2016 Aspect-based Sentiment Analysis task \cite{SemEval2016:task5} for each domain-language pair (\texttt{Best}) as well as against the two best single models of the competition: \texttt{IIT-TUDA} \cite{Kumar2016}, which uses large sentiment lexicons for every language, and \texttt{XRCE} \cite{Brun2016}, which uses a parser augmented with hand-crafted, domain-specific rules. In order to ascertain that the hierarchical nature of our model is the deciding factor, we additionally compare against the sentence-level convolutional neural network of Ruder et al. \shortcite{Ruder} (\texttt{CNN}) and against a sentence-level Bi-LSTM (\texttt{LSTM}), which is identical to the first layer of our model.\footnote{To ensure that the additional parameters do not account for the difference, we increase the number of layers and dimensions of \texttt{LSTM}, which does not impact the results.}

\section{Results and Discussion}

We present our results in Table \ref{tab:results}. Our hierarchical model achieves results superior to the sentence-level CNN and the sentence-level Bi-LSTM baselines for almost all domain-language pairs by taking the structure of the review into account. We highlight examples where this improves predictions in Table \ref{tab:predictions}.

In addition, our model shows results competitive with the best single models of the competition, while requiring no expensive hand-crafted features or external resources, thereby demonstrating its language and domain independence. Overall, our model compares favorably to the state-of-the-art, particularly for low-resource languages, where few hand-engineered features are available. It outperforms the state-of-the-art on four and five datasets using randomly initialized and pre-trained embeddings respectively.

\begin{table}[]
\centering
\begin{tabular}{l l c c}
\textbf{Id} & \textbf{Sentence} & \textbf{LSTM} & \textbf{H-LSTM} \\\hline
1.1 & No Comparison & \textit{negative} & \textit{positive} \\
\multirow{2}{*}{1.2} & It has great sushi and & \multirow{2}{*}{\textit{positive}} & \multirow{2}{*}{\textit{positive}} \\
& even better service. &  & \\\hline
\multirow{2}{*}{2.1} & Green Tea creme & \multirow{2}{*}{\textit{positive}} & \multirow{2}{*}{\textit{positive}} \\
& brulee is a must! &  &  \\
\multirow{2}{*}{2.2} & Don't leave the & \multirow{2}{*}{\textit{negative}} & \multirow{2}{*}{\textit{positive}} \\
& restaurant without it. &  & 
\end{tabular}
\caption{Example sentences where knowledge of other sentences in the review (not necessarily neighbors) helps to disambiguate the sentiment of the sentence in question. For the aspect in 1.1, the sentence-level LSTM predicts \textit{negative}, while the context of the service and food quality in 1.2 allows the H-LSTM to predict \textit{positive}. Similarly, for the aspect in 2.2, knowledge of the quality of the green tea crème brulée helps the H-LSTM to predict the correct sentiment.}
\label{tab:predictions}
\end{table}

\subsection{Pre-trained embeddings}

In line with past research \cite{Collobert2011a}, we observe significant gains when initializing our word vectors with pre-trained embeddings across almost all languages. Pre-trained embeddings improve our model's performance for all languages except Russian, Arabic, and Chinese and help it achieve state-of-the-art in the Dutch phones domain. We release our pre-trained multilingual embeddings so that they may facilitate future research in multilingual sentiment analysis and text classification\footnote{\url{https://s3.amazonaws.com/aylien-main/data/multilingual-embeddings/index.html}}.

\subsection{Leveraging additional information}

As annotation is expensive in many real-world applications, learning from only few examples is important. Our model was designed with this goal in mind and is able to extract additional information inherent in the training data. By leveraging the structure of the review, our model is able to inform and improve its sentiment predictions as evidenced in Table \ref{tab:predictions}.

The large performance differential to the state-of-the-art for the Turkish dataset where only 1104 sentences are available for training and the performance gaps for high-resource languages such as English, Spanish, and French, however, indicate the limits of an approach such as ours that only uses data available at training time.

While using pre-trained word embeddings is an effective way to mitigate this deficit, for high-resource languages, solely leveraging unsupervised language information is not enough to perform on-par with approaches that make use of large external resources \cite{Kumar2016} and meticulously hand-crafted features \cite{Brun2016}.

Sentiment lexicons are a popular way to inject additional information into models for sentiment analysis. We experimented with using sentiment lexicons by Kumar et al. \shortcite{Kumar2016} but were not able to significantly improve upon our results with pre-trained embeddings\footnote{We tried bucketing and embedding of sentiment scores as well as filtering and pooling as in \cite{Vo2015}}. In light of the diversity of domains in the context of aspect-based sentiment analysis and many other applications, domain-specific lexicons \cite{Hamilton2016} are often preferred. Finding better ways to incorporate such domain-specific resources into models as well as methods to inject other forms of domain information, e.g. by constraining them with rules \cite{Hu2016} is thus an important research avenue, which we leave for future work.

\section{Conclusion}

In this paper, we have presented a hierarchical model of reviews for aspect-based sentiment analysis. We demonstrate that by allowing the model to take into account the structure of the review and the sentential context for its predictions, it is able to outperform models that only rely on sentence information and achieves performance competitive with models that leverage large external resources and hand-engineered features. Our model achieves state-of-the-art results on 5 out of 11 datasets for aspect-based sentiment analysis.

\section*{Acknowledgments}

We thank the anonymous reviewers, Nicolas Pécheux, and Hugo Larochelle for their constructive feedback.
This publication has emanated from research conducted with the financial support of the Irish Research Council (IRC) under Grant Number EBPPG/2014/30 and with Aylien Ltd. as Enterprise Partner as well as from research supported by a research grant from Science Foundation Ireland (SFI) under Grant Number SFI/12/RC/2289.

\bibliography{hierarchical_absa}
\bibliographystyle{emnlp2016}

\end{document}